\documentclass[runningheads]{llncs}

\usepackage[T1]{fontenc}
\usepackage{graphicx}
\usepackage{hyperref}
\usepackage{amssymb}
\usepackage{amsmath}
\usepackage{booktabs} 
\usepackage{color}
\usepackage{wrapfig}
\usepackage{marvosym}
\begin{document}
% \title{Towards Realistic Medical Shape Generation: Smooth Manifold Learning on Synchronized Spectral Embeddings}
\title{ToothForge: Automatic Dental Shape Generation using Synchronized Spectral Embeddings}
\titlerunning{Dental Shape Generation using Synchronized Spectral Embeddings}

\author{Tibor Kubík\inst{1,2(\textrm{\Letter})
%,\thanks{Corresponding author: T. Kubík. Email: \email{ikubik@fit.vut.cz}}
} \and
François Guibault\inst{1}\and
Michal Španěl\inst{2} \and
Hervé Lombaert\inst{1}}
%index{Kubík, Tibor}
%index{Guibault, François}
%index{Španěl, Michal}
%index{Lombaert, Hervé}

\authorrunning{T. Kubík et al.}
\institute{
Polytechnique Montréal, Montréal, Canada
\and
Brno University of Technology, Brno, Czech Republic\\
\email{ikubik@fit.vut.cz}
}
\maketitle

% Spectral methods have proven effective in many medical domains, but remain underexplored in dental applications.
\begin{abstract}
We introduce \emph{ToothForge}, a spectral approach for automatically generating novel 3D teeth, effectively addressing the sparsity of dental shape datasets. 
By operating in the spectral domain, our method enables compact machine learning modeling, allowing the generation of high-resolution tooth meshes in milliseconds.
However, generating shape spectra comes with the instability of the decomposed harmonics.
To address this, we propose modeling the latent manifold on \emph{synchronized} frequential embeddings.
Spectra of all data samples are aligned to a common basis prior to the training procedure, effectively eliminating biases introduced by the decomposition instability.
Furthermore, synchronized modeling removes the limiting factor imposed by previous methods, which require all shapes to share a common fixed connectivity.
Using a private dataset of real dental crowns, we observe a greater reconstruction quality of the synthetized shapes, exceeding those of models trained on unaligned embeddings.
We also explore additional applications of spectral analysis in digital dentistry, such as shape compression and interpolation.
ToothForge facilitates a range of approaches at the intersection of spectral analysis and machine learning, with fewer restrictions on mesh structure. 
This makes it applicable for shape analysis not only in dentistry, but also in broader medical applications, where guaranteeing consistent connectivity across shapes from various clinics is unrealistic.
The code is available at \url{https://github.com/tiborkubik/toothForge}.
% Using a variational autoencoder, we learn a smooth latent manifold from which new data can be sampled, decoded, and reconstructed into the spatial domain. This enables the generation of realistic 3D shapes while maintaining computational efficiency. We validate our method on a dataset of 3D teeth shapes, demonstrating its ability to produce high-quality synthetic samples closely resembling real data.

\end{abstract}

%%%%%%%%%%%%%%%%%%%%%%%%%%%%%%%%%%%%%%%%%%%
%
%
%
%%%%%%%%%%%%%%%%%%%%%%%%%%%%%%%%%%%%%%%%%%%
\section{Introduction}
\label{sec:introduction}
3D representations provide important insight into the analysis of anatomical shapes. In digital dentistry, they enable advances in diagnosis, treatment planning, and patient-specific prosthetics design. For tasks within this domain, such as automatic crown design, data-driven approaches have been proposed~\cite{golriz:crown-generation,tian:crown-generation,yang:crown-generation}, offering automated solutions to improve the efficiency of lab technicians.

However, dental shape datasets are typically limited in size due to reasons such as privacy concerns and expensive annotations.
In addition, they are imbalanced due to anatomical factors.
For example, third molars are less commonly represented due to their frequent extraction in clinical practice.
These persistent real-world challenges are unlikely to change, hindering the potential of data-driven tasks that rely on rich datasets to train robust models.
Synthetic data generation has emerged as a solution to address data scarcity and class imbalance in 2D domain~\cite{bronstein2021geometricdeeplearninggrids}.
Our work explores synthetic shape generation to enhance the performance of learning-based analysis of 3D digital dentistry, motivated and resolved by what follows.

\subsubsection{Tackling shape dataset scarcity is challenging.}
Although synthetic data generation has been demonstrated in the 2D medical image domain~\cite{fuente:synthetizing-in-images}, generating synthetic data for 3D shapes introduces unique difficulties. 
Non-Euclidean 3D shapes require intricate geometric operations without the benefit of an underlying regular grid. 
Maintaining spatial relationships further increases computational complexity~\cite{bronstein2021geometricdeeplearninggrids}.
The requirement of generating new samples \emph{on-the-fly} adds to the challenge, particularly for 3D shape data. 

\subsubsection{Choosing the appropriate 3D represention is key.}
The choice of data representation plays a critical role in designing machine learning models for shape analysis.
A diverse range of shape representations has been explored, including projection-based methods~\cite{LE2017103,su:mvcnn} and 3D grid approaches~\cite{qi:vox,wang:ocnn}. 
Although projection-based techniques allow for the adoption of image-domain architectures, they may introduce ambiguities in selecting optimal projections. 
Volumetric methods enable straightforward 3D convolution designs but face high memory demands.
Other approaches operate directly on the mesh structure~\cite{Babiloni_2023_ICCV,hanocka:meshcnn,hu:subdivnet}, which preserves connectivity information but requires subsampling that leads to the loss of fine anatomical details.
This is a critical problem for accurate medical shape analysis.
Point cloud representations are a compelling alternative.
They eliminate the need for connectivity information, relying solely on point-wise features~\cite{qi:pointnet,qi:pointnet++,wu2024pointtransformerv3simpler,zhao2021pointtransformer}.
This representation has shown effectiveness in various applications, including 3D generative modeling~\cite{achlioptas2018learningrepresentationsgenerativemodels} and the development of anatomical statistical shape models~\cite{adams2024pointssm}.
However, operating in the spectral domain enables shape analysis to operate in the frequential domain, unlocking more compact and efficient approaches when analyzing complex shapes~\cite{reuter:shapedna,styner:spharm-pdm}.

\subsubsection{The key lies in shape spectra.}
Spectral coefficients 
% and spherical harmonics 
encode a shape geometry through its intrinsic properties, often requiring only a limited set of harmonics to capture key features effectively. 
Additionally, spectral representations allow for truncating higher-frequency components as shown in Fig.~\ref{fig:spec-basics}. 
This reduces data dimensionality while retaining essential characteristics, making learning tasks computationally more efficient.
These benefits contrast with spatial representations, which often scale inefficiently with resolution and lose significant information during stratification.
What is more, spectral representations are inherently ordered, 
%(thanks to eigenvalues and degrees and orders of Legendre polynomials in spherical harmonics basis functions), 
reducing the need for expensive neighbor searches in high-dimensional feature spaces.
The effectiveness of spectral approaches has been demonstrated in various learning tasks~\cite{agus2020wish,GOPINATH2019297,ha2022spharm,marin:spec-nn}.
These harmonics are however inherently unstable, and training on such coefficients introduces unwanted distortions into the network. 
We address this issue by using synchronized coefficients during training to eliminate the bias.
We also explore spectral analysis tools in dental imaging for tasks such as tooth compression.

\subsubsection{Using spectral generative models on medical shapes is challenging.}
Although prior work on spectral shape analysis exists~\cite{reuter:shapedna}, the concept of modeling latent spaces based on spectral coefficients for generating novel shapes was first introduced in~\cite{LEMEUNIER2022131}.
% The idea of modeling latent spaces on spectral coefficients for generating novel shapes was introduced in~\cite{LEMEUNIER2022131}.
Here, the authors trained a spectral autoencoder (AE) on a dataset of human poses and showed promising results in shape interpolation. 
However, its application to medical shape data remains challenging for at least two reasons. 
Firstly, AEs lack the regularization required to ensure a smooth latent space, which is particularly important for small-scale datasets such as those in medical domains.
Without this, sampling in latent space voids yields implausible reconstructions. 
Our model design choice is $\beta$-VAE with cyclical annealing schedule~\cite{fu:2019-cyclical}.
We chose a $\beta$-VAE over an AE or a standard VAE for its ability to balance reconstruction accuracy and feature disentanglement. 
This allows better control over the trade-off between geometric fidelity and a smooth latent space, which is essential for generating plausible shapes. 
More complex models such as GANs or diffusion models were avoided due to their data-intensive nature and training instability, making them less suitable for low-scale datasets.

More critically, the framework in~\cite{LEMEUNIER2022131} and its follow-up work~\cite{LEMEUNIER2023191} requires fixed connectivity of all training meshes. 
This makes it unapplicable in real-world medical scenarios, where shapes scanned at different clinics using various intra-oral scanners have different connectivity.
To address this, we introduce \emph{spectral synchronization} during training, aligning all shape spectra to a common base.
This alignment ensures that the model learns from spectral coefficients in a shared spectral space, regardless of connectivity. 
It also minimizes the bias caused by the instability of harmonics, producing more reliable predictions.
%Spectral synchronization is required only during training. 
%During synthesis, the decoder generates aligned coefficients, which are then projected into the spatial domain through a common basis by a single sparse matrix multiplication. 
%This process minimizes computational overhead during deployment, making it practical for real-world applications.

\subsubsection{Our contributions in spectral shape learning and digital dentistry.}
\label{subsec:contributions}
We introduce an efficient and accurate approach to real-time generation of synthetic tooth shapes utilising spectral analysis. 
Beyond exploring low-pass filtering for tooth crowns, spectral alignment, and potentionally correspondence, we are the first to utilize modal coefficients to train a spectral autoencoder in digital dentistry. 
A key contribution of our work is the use of synchronized embeddings during training, a novel approach that eliminates noise arising from the inherent instability of harmonics. 
This advancement also enables the use of datasets with varying mesh connectivity, advancing the state-of-the-art in this field.
Another important outcome of this solution, which arises naturally from training on synchronized frequency coefficients, is the vertex-wise correspondence among all generated shapes.
We show how critical design choices ensure both high fidelity in the generated shapes and smoothness in the latent space. 
Comparative benchmarks highlight the superiority of our method over spatial approaches.

%%%%%%%%%%%%%%%%%%%%%%%%%%%%%%%%%%%%%%%%%%%
%
%
%
%%%%%%%%%%%%%%%%%%%%%%%%%%%%%%%%%%%%%%%%%%%
\section{Methodology}
\label{sec:methodology}
We first outline the principles of differential operators and spectral analysis on manifolds.
Building on this, we discuss an approach that utilizes spectral coefficients as features for generative model training. 
We address limitations inherent in this method by introducing the concept of spectral synchronization and latent space regularization.
This enhances the versatility of the framework in real-world medical scenarios such as tooth generation.

\subsection{Spectral Decomposition on Teeth Shapes}
\label{subsec:spectral-decomposition-on-teeth-shapes}

\subsubsection{Discrete manifolds.} In the discrete setting, a manifold $\mathcal{X}$ is sampled at $n$ points in $\mathbb{R}^3$, and its approximation is given by a triangular mesh $(\mathcal{V}, \mathcal{E}, \mathcal{F})$, where $\mathcal{V} = \{v, \cdots, v_n\}$ is a set of the sampled points $v = (x, y, z) \in \mathbb{R}^3$, $\forall v \in \mathcal{V}$, edges $\mathcal{E}\subseteq \mathcal{V}^2$ and faces $\mathcal{F}\subseteq \mathcal{V}^3$.
Based on linear FEM, the discretization of the Laplace-Beltrami operator $\Delta_\mathcal{X} f$ takes the form of an $n \times n$ sparse matrix $\mathbf{L} = \mathbf{A}^{-1}\mathbf{W}$.
Here, the \emph{mass matrix}~$\mathbf{A}$ is a sparse diagonal matrix of elements $a_i~=~\frac{1}{3}\sum_{jk:(ijk) \in \mathcal{F}} A_{i,j,k}$ where $A_{i,j,k}$ denotes the triangle area formed by vertices $i$, $j$ and $k$. $\mathbf{W}$ is a symmetric matrix of edge-wise weights~(equivalent to classical \emph{cotangent formula}~\cite{meyer:diff-geom-2003}):

\begin{equation}
    \textbf{W}_{ij} = 
    \begin{cases}
      -(cot\alpha_{ij} + cot\beta_{ij}) & i \neq j, v_j \in N_1(v_i), \\
      \sum_{v_j \in N_1(v_i)}(cot\alpha_{ij} + cot\beta_{ij}) & i = j, \\
      0 & v_j \notin N_1(v_i),
    \end{cases}
    \tag{1}
\end{equation}
where $N_1(v_i)$ is the 1-ring neighborhood of the vertex $v_i \in \mathcal{V}$, and $\alpha_{ij}$ and $\beta_{ij}$ are angles opposite to the edge formed by vertices $v_i \in \mathcal{V}$ and $v_j \in \mathcal{V}$.

\subsubsection{Spectral analysis on meshes.} In Riemannian geometry, the orthogonal eigenbasis of the Laplacian obtained by the eigendecomposition process is used to define an analogy of the Fourier transform.
In matrix notation, it is defined as a square matrix $\mathbf{L} = \mathbf{\Phi}\mathbf{\Lambda}\mathbf{\Phi}^T$. 
Here, $\mathbf{\Lambda}$ is a diagonal matrix of real, non-negative eigenvalues $\lambda_i \in \mathbb{R}$, $\mathbf{\Lambda} = \text{diag}(\lambda_1, \dots, \lambda_n)$, where $\lambda_1 = 0 \leq \dots \leq \lambda_n$. 
% Throughout the rest of the paper, eigenvalues are always assumed to be increasingly ordered.
Note that the eigenvalues are increasingly ordered due to the inverse of $\mathbf{A}$ in the Laplacian calculation.
$\mathbf{\Phi}$ is a matrix of the corresponding eigenvectors $\mathbf{\Phi} = (\phi_i, \dots, \phi_n)$, such that $\mathbf{L}\phi_{i} = \lambda_i\phi_i$, $\forall i = 1, \dots, n$, $\phi_i \in \mathbb{R}^n$. 
The low frequencies correspond to the smallest eigenvalues, encoding coarse shape information (such as crown length of incisors). 
Large eigenvalues correspond to high-frequency shape information such as molar cusp morphologies, exhibiting rapid oscillations. 
In addition, any function defined at the vertices of a mesh can be represented as:
\begin{equation}
    g = \sum_{i = 1}^{n} \langle f, \phi_i \rangle \phi_i = \mathbf{\Phi}\mathbf{\Phi}^T\!f.
    \tag{2}
\end{equation}
Here, $\langle f, \phi_i \rangle$ represent \emph{spectral coefficients}, frequency coordinates that contain information about the geometry of the original vertices in a compressed way.

\begin{figure}[t!]
\centering
\includegraphics[width=\textwidth]{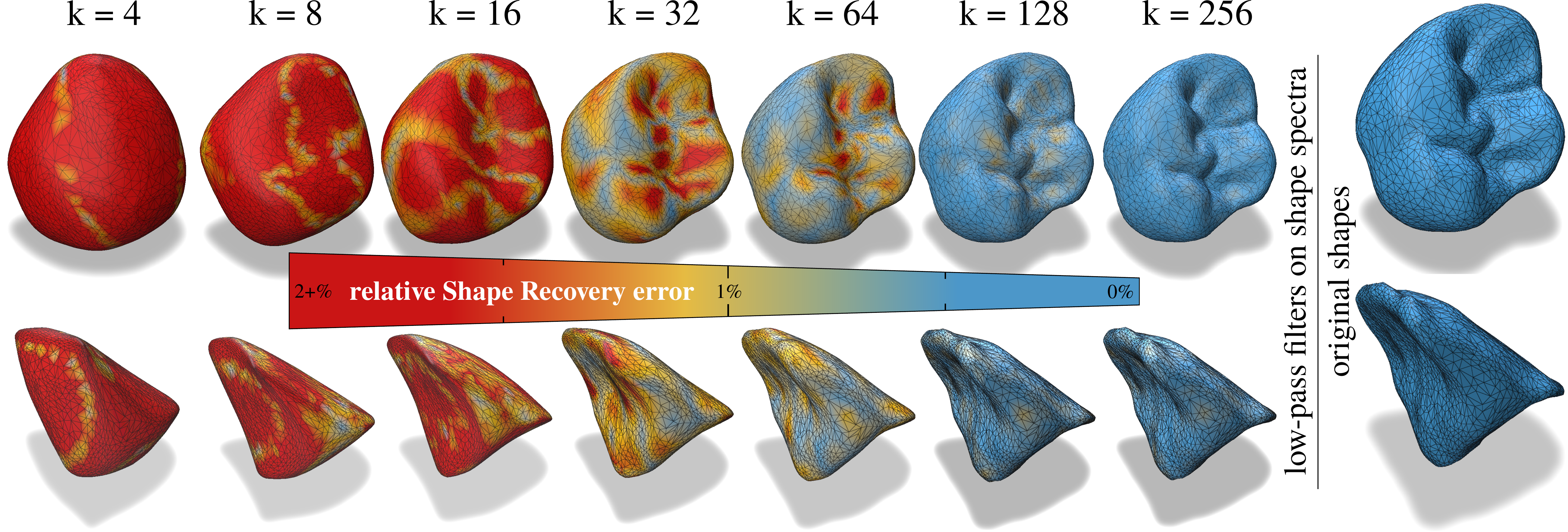}
\caption{\textbf{Example of low-pass filtering on molar and incisor shapes.}
Higher spectral components preserve finer geometric details.
With as low as 128 coefficients, the recovered shapes preserve most of the anatomical details.}
\label{fig:spec-basics}
\end{figure}

A low-dimensional representation of a mesh can be obtained by keeping only the leading spectral coefficients. 
We consider all mesh vertices as a matrix $\mathbf{V} \in \mathbb{R}^{n\times3}$ whose columns define their $x$, $y$ and $z$ positions in space.
Given $\mathbf{V}$, we can reconstruct an approximate version of the mesh by using the $k$-truncated eigenbasis:
\begin{equation}
    \mathbf{V}^k = \sum_{i = 1}^{k} \langle \mathbf{V}, \phi_i \rangle \phi_i = \mathbf{\Phi}_k\mathbf{\Phi}_k^T\!\mathbf{V}.
    \tag{3}
\end{equation}
To quantify the information loss, one can measure the \emph{Shape Recovery (SR) error} defined as the mean squared error between vertex positions of the original and smoothed shape.
Examples of tooth approximations are provided in Fig.~\ref{fig:spec-basics}.

\subsection{Spectral Autoencoding: the Baseline Approach}
\label{subsec:spectral-autoencoders}
\emph{Spectral Autoencoder}~(\emph{SAE-LP-k}), a framework originally proposed in~\cite{LEMEUNIER2022131}, utilizes the concepts from spectral analysis.
SAE-LP-k processes 3D meshes by first transforming their spatial vertices into spectral coefficients. 
This is achieved by projecting each vertex onto its corresponding eigenspace using a truncated set of $k$ eigenvectors, thus forming $k$-banded spectral coefficients. 
The computation of these coefficients serves as input to the autoencoder. 
Mathematically, the spectral coefficients $\mathbf{C}_k$ using $k$-truncated eigenbasis $\mathbf{\Phi} \in \mathbb{R}^{n \times k}$ for a mesh with $n$ vertices are computed as:
\begin{equation}
\mathbf{C}_k = \mathbf{\Phi}^T_k \mathbf{V},
\tag{4}
\end{equation}

\noindent where $\mathbf{V} \in \mathbb{R}^{n \times 3}$ represents the spatial coordinates of the mesh vertices.
This approach addresses issues such as the high number and disordered arrangement of vertices, making the network invariant to rotation, scale, and translation.

Spectral autoencoder function consists of an encoder $e_\theta(z | \mathbf{C}_k)$ and a decoder $d_\gamma(\mathbf{\hat{C}}_k | z)$.
$e_\theta: \mathbb{R}^{k \times 3} \to \mathbb{R}^d$ maps the spectral coefficients to a single deterministic point in $d$-dimensional latent space, and $d_\gamma: \mathbb{R}^d \to \mathbb{R}^{k \times 3}$ reconstructs the coefficients back to the spectral domain.
The reconstruction loss $\mathcal{L}_{\text{rec}} = \|\mathbf{C}_k - \mathbf{\hat{C}}_k\|^2$ ensures fidelity to the original representation within the spectral domain, without reverting to the spatial domain.
The parameters of the encoder and decoder are both computed by convolutional neural networks: since the array of spectral coefficients is already ordered, the network is able to learn meaningful features by sliding a convolution kernel over coefficients.
The learnable (un)pooling procedure is also simple and resembles classical (un)pooling operations.

\subsection{Eliminating Constant Connectivity Constraints via Spectral Synchronization}
\label{subsec:spectral-synchronization}
While the spectral autoencoder effectively learns a latent representation from modal coefficients, it operates under the restrictive assumption that all shapes in the training dataset have identical connectivity.
This assumption allows for a direct computation of spectral coefficients using a common $k$-truncated eigenbase $\mathbf{\Phi}_{k\text{-ref}}$ since the per-vertex correspondence is well defined among all shapes within the dataset.
However, this requirement is unrealistic for medical datasets, where meshes exhibit variable connectivities. 
Moreover, since the exact spectrum is rarely known and is usually approximated numerically, this introduces several issues, leading to incompatible bases between meshes.
\emph{Sign flips} can occur, as all scalar multiples of an eigenfunction are contained in the same eigenspace.
Arbitrary basis could be constructed of \emph{higher dimensional eigenspaces}.
\emph{Switching of eigenfunctions} can occur, where two close eigenvalues can switch their order due to numerical instabilities or geometry deformations.

\begin{figure}[t!]
\centering
\includegraphics[width=\textwidth]{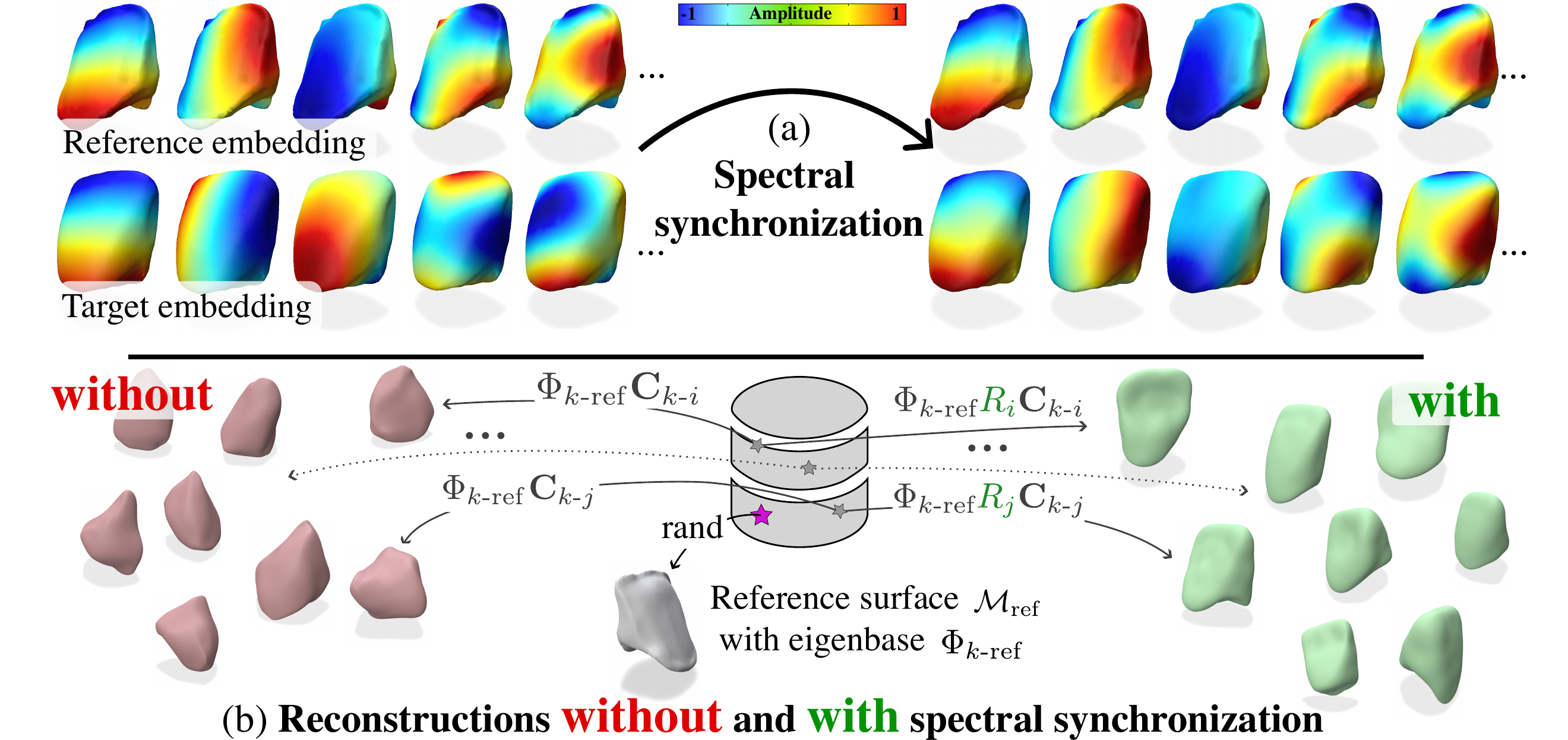}
\caption{\textbf{Impact of spectral synchronization.}
We perform spectral synchronization across all shapes in the dataset as depicted in (a). This process standardizes the harmonics, ensuring they all \emph{speak a common language}. As illustrated in (b), ommiting this step leads to incorrect reconstructions, introducing significant noise into the network when trained on unaligned data.}
\label{fig:spec-sync}
\end{figure}

We utilize a robust computation of $\mathbf{\Phi}_{k\text{-ref}}$ to develop a method that remains invariant to connectivity and vertex number differences among shapes within the dataset, and that suppresses any biases caused by instable computation of harmonics.
Given a dataset of meshes $\{\mathcal{M}_i\}_{i=1}^{N}$ each with its respective $k$-truncated basis $\mathbf{\Phi}_{k\text{-}i}$ and coefficents $\mathbf{C}_{k\text{-}i}$, we randomly select a mesh $\mathcal{M}_{\text{ref}}$ and compute its truncated basis $\mathbf{\Phi}_{k\text{-ref}} \in \mathbb{R}^{n \times k}$. We assume the existence of a spectral transformation $R_i$ with dimensionality of $k \times k$ for each shape $\mathcal{M}_i$:

\begin{equation}
    R_i = ((\mathbf{\Phi}_{k\text{-ref}})^T\mathbf{\Phi}_{k\text{-ref}})^{-1} (\mathbf{\Phi}_{k\text{-ref}})^T\mathbf{\Phi}_i c_i, \quad \forall i \in \{1, \dots, N\},
    \tag{5}
\end{equation}

\noindent where $c_i$ is an unknown correspondence map that matches the rows of $\mathbf{\Phi}_{k\text{-ref}}$ with the equivalent rows of $\mathbf{\Phi}_i$. 
The transformation relies on identifying the correspondence map $c_i$.
This map is optimized by minimizing the $L_2$-norm difference between the aligned spectral coefficients:

\begin{equation}
    \|\mathbf{\Phi}_{k\text{-i}}c_i\mathbf{C}_i - \mathbf{\Phi}_{k\text{-ref}}R_i\mathbf{C}_i\|^2,
    \tag{6}
\end{equation}

\noindent where $\mathbf{C}_i$ is the spectral coefficient array of $\mathcal{M}_i$.
Symmetry is enforced by adding the inverse mapping.
After alignment, the spectral coefficients of all shapes are represented in the common reference basis, enabling consistent downstream processing. 
Additionally, $c_i$ provides vertex-wise correspondence between each shape and the reference.
The bias introduced during training without synchronization is illustrated in Fig.~\ref{fig:spec-sync}.

\begin{figure}[t!]
\centering
\includegraphics[width=\textwidth]{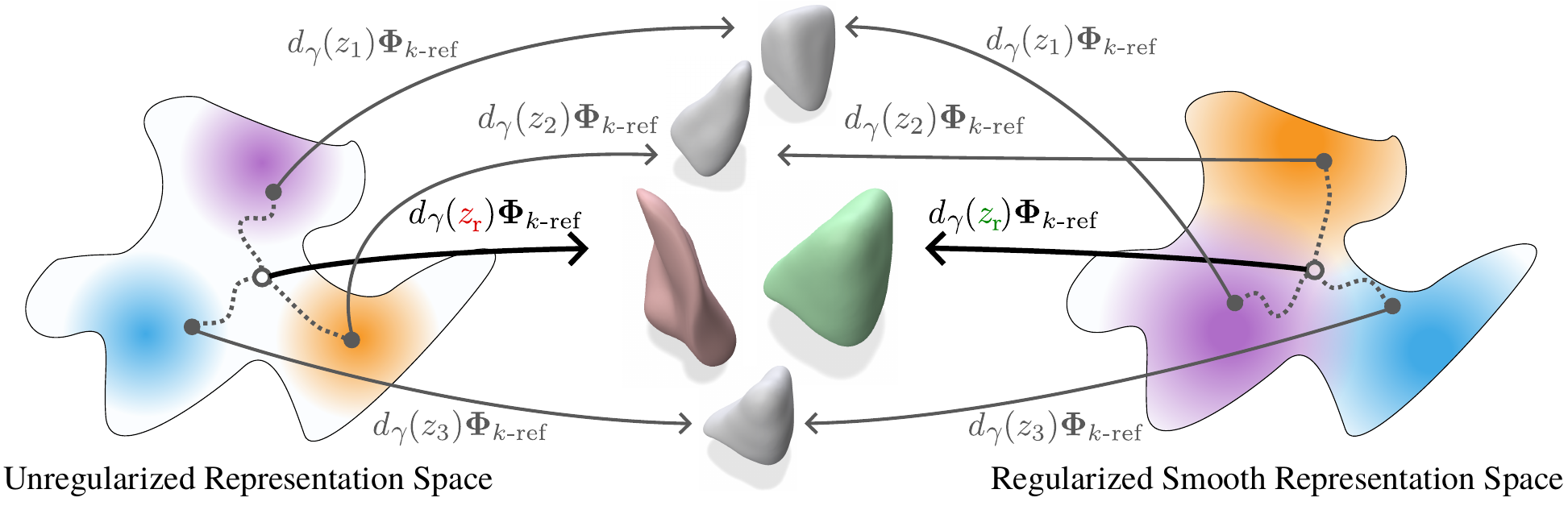}
\caption{\textbf{Effect on regularization of latent space modeled on synchronized spectral coefficients.} 
The process denoted by the solid arrows $d_\gamma(z)\mathbf{\Phi}_{k\text{-ref}}$ involves decoding the latent vector and projecting the result through a common basis into the spatial domain. 
If $z$ is sampled from a void region of the manifold, the resulting geometry becomes corrupted. 
Regularization reduces the likelihood of such cases, ensuring more reliable outputs.}
\label{fig:regularization}
\end{figure}

\subsection{Smooth Latent Manifold Design on Synchronized Spectral Coefficients with $\beta$-VAEs}
\label{subsec:uniform-distribution-disentangled-representations}
While (synchronized) spectral autoencoders offer a powerful means of compressing and reconstructing spectral coefficients of 3D shapes, their use for generating novel samples given a sparse initial dataset is limited.
When working with low-scale medical shape datasets, it becomes especially challenging to model a smooth latent manifold.
Small datasets lack the diversity needed to cover the full space of possible shapes, leading to \emph{gaps} in the learned latent space.

We modify the encoder function $e_\theta$ defined in Sec.~\ref{subsec:spectral-autoencoders} by introducing a stochastic component, mapping synchronized spectral coefficients into vectors of mean $\mu$ and variance $\Sigma$, representing a distribution: $e_\theta(z | R\textbf{C}_k) \sim \mathcal{N}(\mu, \Sigma)$.
Notice that the coefficients are aligned with their corresponding spectral alignment matrix so all shapes are represented on a common basis.
The decoder $d_\gamma(\mathbf{\hat{C}}_k | z)$ remains unchanged, but since the network is trained on aligned spectra, the reconstructed coefficients $\mathbf{\hat{C}}_k$ follow the same trend.
The objective is a combination of the reconstruction loss and $\beta$-weighted Kullback-Leibler (KL) divergence, balancing between reconstruction fidelity and distribution approximation:

\begin{equation}
    \mathcal{L} = \|R\mathbf{C}_k - \mathbf{\hat{C}}_k\|^2 + \beta \text{KL}(e_\theta(z | R\textbf{C}_k) | p(z)).
    \tag{7}
\end{equation}

\noindent Positive values of $\beta$ puts pressure on the bottleneck to match the prior $p(z)$. 
%Incorporating disentanglement properties comes at the cost of diminished reconstruction quality of $\mathbf{\hat{C}}_k$.
The effect of this regularization on teeth spectral generation is visualized in Fig.~\ref{fig:regularization}.

\begin{figure}[t!]
\centering
\includegraphics[width=\textwidth]{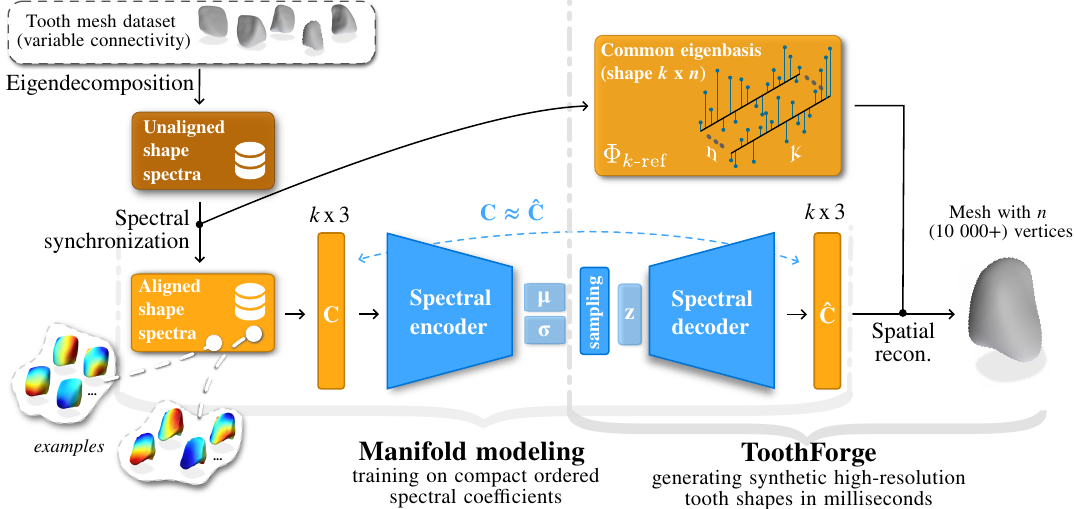}
\caption{\textbf{Method outline.} Our framework for generating synthetic shapes of teeth via sampling on a latent manifold.
Such manifold is modeled using spectral \emph{synchronized} (spectrally aligned to common basis) coefficients of tooth shapes, denoted as $C$.
For novel data sampling using ToothForge, two ingredients are necessary: decoder weights for infering novel modal coefficient $\hat{C}$ and common eigenbasis $\mathbf{\Phi}_{k\text{-ref}}$ to project it to spatial domain. 
Synthetising novel tooth shape with $n$ ($10\,000+$) vertices is performed in order of miliseconds thanks to operating in the spectral domain.}
\label{fig:method-outline}
\end{figure}

Post-training, two ingredients are necessary to generate new data: trained decoder weights $\phi$ and common eigenbase $\mathbf{\Phi}_{k\text{-ref}}$.
Novel vertex matrix $\mathbf{V}_{\text{syn}} = d_\gamma(z_r)\mathbf{\Phi}_{k\text{-ref}} = \mathbf{\hat{C}}_k\mathbf{\Phi}_{k\text{-ref}}$ is coupled with the edges $\mathcal{E}$ of the reference mesh $\mathcal{M}_{\text{ref}}$ to form a realistic high-resolution synthetic mesh.
The visual summary of the whole framework can be seen in Fig.~\ref{fig:method-outline}.

%%%%%%%%%%%%%%%%%%%%%%%%%%%%%%%%%%%%%%%%%%%
%
%
%
%%%%%%%%%%%%%%%%%%%%%%%%%%%%%%%%%%%%%%%%%%%
\section{Experimental Setup}
\label{sec:experimental-setup}

\subsubsection{Dataset.}
\label{subsec:dataset}
We conducted our experiments on a private dataset of 430 tooth shapes provided by industrial partner, approved by an ethics committee.
It is divided into three categories: incisors, premolars and molars. 
The canine class was excluded from the experiments due to insufficient data availability.
Separate models were trained for each category.
The samples are fully anonymized, containing only the mesh data without additional information.
Data were divided by 80-20 ratio into training/validation and testing.

\subsubsection{Evaluation metrics.}
\label{subsec:metrics}
To compare two ordered coefficient sets of equal size, $C_1 \subseteq \mathbb{R}^3 $ and $C_2 \subseteq \mathbb{R}^3$, we calculate the distance as $d_{\text{MSE}}(C_1, C_2) = \|C_1 - C_2\|^2$.
For unordered vertex sets $S_1 \subseteq \mathbb{R}^3$ and $S_2 \subseteq \mathbb{R}^3$ we use the \emph{Chamfer distance} defined as
$d_{\text{CD}}(S_1, S_2)~=~\sum_{x \in S_1} \min_{y \in S_2} \| x - y \|_2^2 + \sum_{y \in S_2} \min_{x \in S_1} \| x - y \|_2^2$.
To evaluate how well the distribution of synthesized samples $P$ approximates the real distribution $G$, we employ a \emph{Minimum Matching Distance} (\emph{MMD}) metric, calculated using either $d_{\text{MSE}}$ or $d_{\text{CD}}$ as structural distance.
MMD quantifies \emph{fidelity} of $P$ with respect to $G$. 
Each sample in $G$ is matched to the closest sample in $P$, and the average distance is computed.
Lower MMD scores indicate that the samples in $P$ closely resemble those in $G$, reflecting higher fidelity.
%\subsubsection{Coverage (COV).}
%Coverage measures how well $P$ represents $G$. 
%For each sample in $P$, the nearest neighbor in $G$ is identified based on either $d_{\text{MSE}}$ or %$d_{\text{CH}}$.
%The score is reported as the fraction of samples in $G$ that are matched to $P$.
%A higher coverage score suggests that $G$ is well-represented in $P$.

%\subsubsection{Minimum Matching Distance (MMD).}
%MMD quantifies the \emph{fidelity} of $P$ with respect to $G$. 
%Each sample in $G$ is matched to the closest sample in $P$, and the average distance is computed.
%Lower MMD scores indicate that the samples in $P$ closely resemble those in $G$, reflecting higher fidelity.

\subsubsection{Implementation details.}
\label{subsec:implementation-details}

We utilize a compact model with 5-stage encoder and decoder with hidden feature sizes of 32, 64, 128, 256 and 512.
Training was carried out on a single Tesla T4 with 16GB of VRAM spanned over approximately 2 hours. The model was optimized by AdamW optimizer with an initial learning rate of 1e-4, dynamically changed using cosine annealing restarts each $10\,000$ iterations.
The value of $\beta$ in $\beta$-VAE changed within the range of 0 to 0.05 using cyclical annealing scheduler.
We fix the value of $k$ to 256 when generating truncated spectral coefficients.
Latent size is fixed to 16.
%%%%%%%%%%%%%%%%%%%%%%%%%%%%%%%%%%%%%%%%%%%
%
%
%
%%%%%%%%%%%%%%%%%%%%%%%%%%%%%%%%%%%%%%%%%%%
\section{Results}
\label{sec:results}
We report the key quantitative results in Table~\ref{tab:results}.
Reconstruction metrics are the average distance errors computed from all test cases for each tooth class.
To evaluate the quality of synthesized shapes, we randomly sample from $\mathcal{N}(0, I)$, decode, and reconstruct $1\,000$ latent vectors and use training samples as the ground truth distribution.
The measured metrics remain consistently low across all tooth classes, demonstrating the learned representation's ability to generalize to unseen shapes while providing reliable reconstructions.
We also report the average time of generating $1\,000$ novel tooth shapes.
It takes approximately 1 millisecond to perform both spectral decoder inference and projection to the spatial domain, highlighting ToothForge's capability to enhance downstream dental tasks by synthesizing novel shapes without significant additional overhead.
The text follows with qualitative and comparative analysis.

\begin{table}[t!]
    \caption{\textbf{Quantitative evaluation of generated shapes using ToothForge.} Metrics are the average scores across all test cases for given tooth class.
    $\text{MMD}$ is calculated from $1\,000$ randomly sampled and reconstructed latent vectors.
    Measured times include the decoder's forward pass and spatial reconstruction without batching optimizations.}
    \label{tab:results}
    \centering
    \resizebox{\columnwidth}{!}{%
    
    \begin{tabular}{l@{\hskip 0.5in}c@{\hskip 0.15in}c@{\hskip 0.15in}c@{\hskip 0.15in}c}
        \toprule
        Tooth class & $d_{\text{MSE-spectral}}\downarrow$ & $d_{\text{MSE-spatial}}\downarrow$ & $\text{MMD}\downarrow$ &
        Shape generation time (ms)$\downarrow$\\
        \midrule
        Incisors    &   $0.08737$   &   $0.00211$   &   $0.00754$ &   $0.71 \pm 0.05$ ($10\,623$ vertices)  \\
        Premolars   &   $0.09764$   &   $0.00209$   &   $0.00716$ &   $0.75 \pm 0.06$ ($11\,960$ vertices)  \\
        Molars      &   $0.03225$   &   $0.00104$   &   $0.00325$ &   $0.77 \pm 0.03$ ($11\,671$ vertices)  \\
        \bottomrule
    \end{tabular}}
\end{table}

\begin{figure}[h!]
\centering
\includegraphics[width=.9\textwidth]{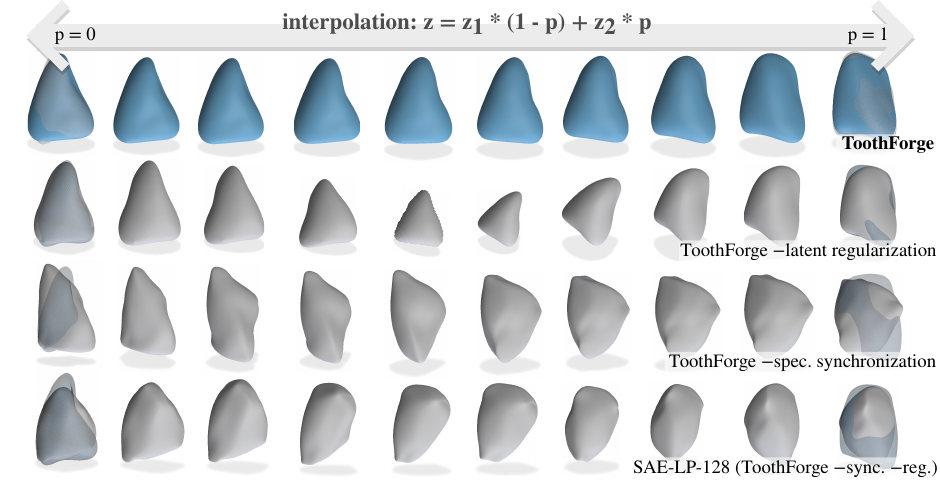}
\caption{\textbf{Synthetic data generation via interpolation in latent space.} 
Individual rows show linear interpolation of latent codes for various framework setups. 
Two incisor samples from the test set were chosen (leftmost and rightmost), encoded into latent representations $z_1$ and $z_2$, sampled along the line $z = z_1 * (1 - p) + z_2 * p$, decoded and finally reconstruted using common basis. 
Ground truth shapes overlay leftmost and rightmost reconstructions.}
\label{fig:res-interp-regularized}
\end{figure}

\subsubsection{Effects of spectral synchronization and latent regularization.} We conduct a qualitative assessment to validate the design choices outlined in Sec.~\ref{subsec:spectral-synchronization} and Sec.~\ref{subsec:uniform-distribution-disentangled-representations} by training on unaligned spectral coefficients and omitting latent space regularization. In the absence of both components, the overall framework design closely resembles SAE-LP-128, as discussed in Sec.~\ref{subsec:spectral-autoencoders} and is currently the state-of-the-art in spectral generative modeling~\cite{LEMEUNIER2022131,LEMEUNIER2023191}.
The results in Fig.~\ref{fig:res-interp-regularized} demonstrate the importance of these components in scenarios with sparse data and various mesh connectivities.
Throughout the interpolation sequence, artifacts from incorrectly reconstructed harmonic coefficients are mitigated, leading to smoother and more realistic shapes while avoiding shrinkage.

\subsubsection{Reconstruction fidelity.} 
\begin{figure}[t!]
\centering
\includegraphics[width=0.9\textwidth]{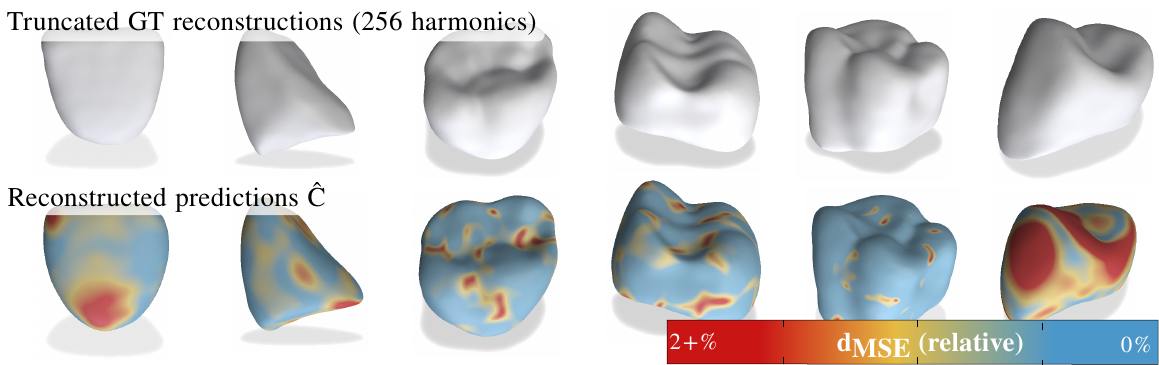}
\caption{\textbf{Reconstructions of unseen tooth shapes.} The predictions accurately capture the overall tooth shape. The reconstructions may appear smoothed at times (rightmost premolar). This is due to inaccuracies in predicting high-frequency components.}
\label{fig:res-recons}
\end{figure}

We demonstrate various reconstruction results in Fig.~\ref{fig:res-recons}.
We use the network to encode and decode the spectral coefficients of the test set.
%and compare the spatial reconstruction with its ground-truth counterparts.
The predictions effectively capture the overall shape of the tooth.
Reconstructions occasionally appear overly smooth, which is attributed to prediction errors in the high-frequency coefficients.

\subsubsection{Spectral vs. spatial: a comparative analysis.} 
\begin{figure}[t!]
\centering
\includegraphics[width=.9\textwidth]{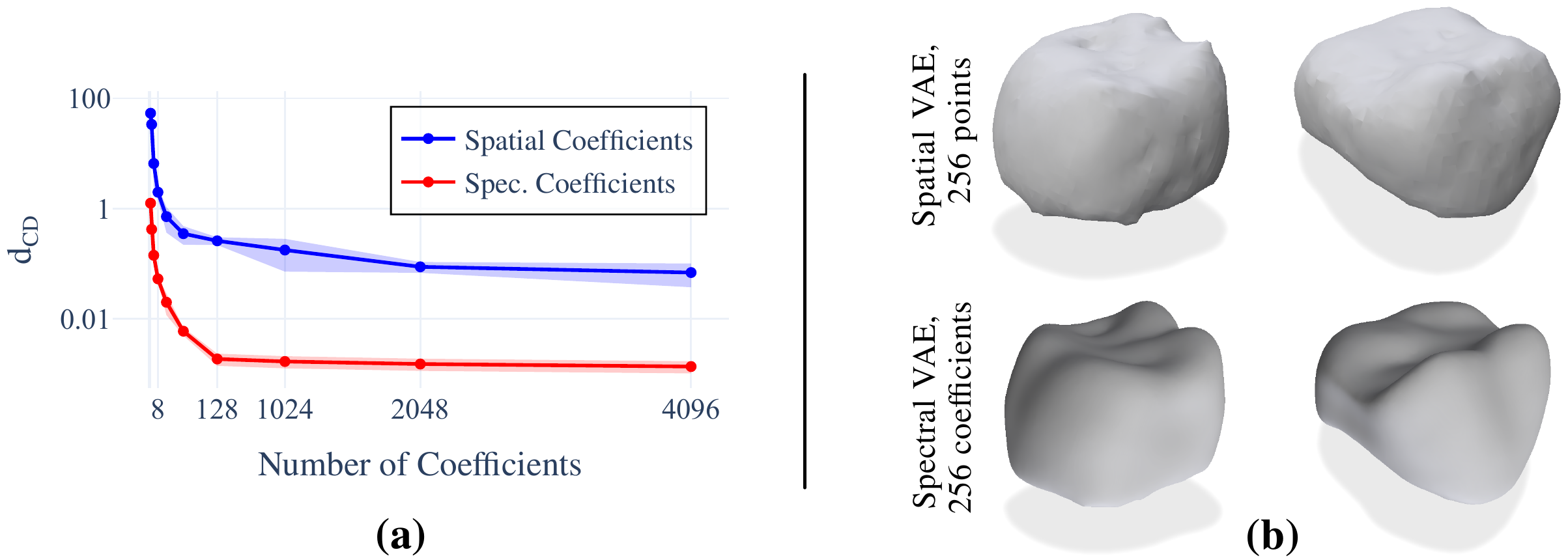}
\caption{\textbf{Spectral vs. spatial coordinates in tooth analysis.} In \textbf{(a)}, Chamfer Distances ($d_{CD}$) for spatial coefficients are computed by sampling a given number of points, reconstructing the mesh using Poisson reconstruction, and then calculating the $d_{CD}$ with the original mesh.
The logarithmic y-axis highlights this difference.
Both methods converge to low error levels as the number of coefficients increases, but the spectral approach remains consistently more efficient. 
Values are scaled by 10.
\textbf{(b)} depicts reconstruction capability of autoencoders when processing 256 coordinates.}
\label{fig:spectral-spatial}
\end{figure}

Spectral coefficients provide a compact representation of 3D data by transforming spatial coordinates into the frequency domain. 
%This dramatically reduces memory usage and accelerates training.
We highlight this compactness for teeth in the graph in Fig.~\ref{fig:spectral-spatial}. 
The graph compares Chamfer distances for spatial and spectral coefficients as a function of the number of coefficients used for reconstruction of all molar cases in our dataset.  
Spectral coefficients exhibit errors lower than that of spatial coefficients, demonstrating their superior compactness in representing geometry.
Spectral coefficients also offer natural ordering that simplifies pooling operations since it allows for hierarchical processing directly along this order, eliminating the need for expensive nearest-neighbor searches in high-dimensional spaces~\cite{qi:pointnet++,wang:dgcnn}.
Reconstructing dental structures with a spectral autoencoder yields smoother but more plausible results, unlike the spatial counterpart, which often adds high-frequency noise and misses key features like cusps.
While smoothing can help eliminate noise from the spatial output, incorporating more anatomical details would require a substantial increase in resolution, making the spectral approach more efficient in on-the-fly generation.

%%%%%%%%%%%%%%%%%%%%%%%%%%%%%%%%%%%%%%%%%%%
%
%
%
%%%%%%%%%%%%%%%%%%%%%%%%%%%%%%%%%%%%%%%%%%%
\section{Conclusion}
\label{sec:conclusion}
In this work, we presented \emph{ToothForge}, a spectral approach to generate novel 3D teeth in real time. 
The main motivation is to address the sparsity of dental shape datasets, so synthesized shapes increase the accuracy of downstream tasks in digital dentistry~\cite{golriz:crown-generation,tian:crown-generation,yang:crown-generation}.
It is important that the method introduces minimal time overhead when used as an augmentation during the training of these tasks.
To address this, we propose a compact autoencoder that learns from spectral embeddings, as mesh spectra are known to provide an effective shape representation \cite{reuter:shapedna}.
However, generating shape spectra comes with the instability of the decomposed harmonics.
We propose modeling the latent manifold on \emph{synchronized} versions of frequential embeddings.
Spectra of all data samples are aligned to a common basis prior to the training procedure, effectively eliminating biases introduced by the decomposition instability.
More importantly, we eliminate the assumption that all shapes in the training dataset must share a common fixed connectivity, a factor imposed by previous works~\cite{LEMEUNIER2022131,LEMEUNIER2023191}. 
Our work extends its application to scenarios where guaranteeing consistent connectivity across shapes is unrealistic.
We evaluated our framework using a private dataset of real dental crowns.
It takes less than a millisecond to generate a new tooth shape with more than $10\,000$ vertices using a single GPU.
Reconstruction accuracy is consistently higher when training on aligned embeddings, supported by qualitative analysis.
Synthetic shapes closely resemble ground truth shapes, achieving the average MMD value of $0.00598$ across tooth classes.
We demonstrate that spatial networks perform poorly compared to spectral frameworks given the same size of input features. 
Using $256$ spectral coefficients is enough to generate realistic teeth, whereas the same number of spatial coefficients generates noisy meshes that lack important anatomical features such as molar cusps. 

In summary, our method, \emph{ToothForge}, provides a tool for synthesizing tooth shapes, introducing a new strategy for data augmentation in dental shape analysis tasks.
This has the potential to significantly enhance their accuracy with minimal computational cost.
Future research will investigate the disentanglement properties of the manifold, specifically how different latent dimensions represent independent and interpretable anatomical features. 
This disentanglement would allow for more precise control over features such as cusp sizes or groove depths in patient-specific crowns. By navigating the disentangled manifold, these anatomical features could be modified in real-time, paving the way for broader applications in the daily practice of dental technicians and beyond the field of dentistry.

%%%%%%%%%%%%%%%%%%%%%%%%%%%%%%%%%%%%%%%%%%%
%
%
%
%%%%%%%%%%%%%%%%%%%%%%%%%%%%%%%%%%%%%%%%%%%
%\begin{credits}
%\subsubsection{\ackname} This study was funded by X (grant number Y).
%
%\subsubsection{\discintname}
%It is now necessary to declare any competing interests or to specifically
%state that the authors have no competing interests. (The authors have no competing interests to declare that %are
%relevant to the content of this article.)
%\end{credits}
\bibliographystyle{splncs04}
\bibliography{paper175}
\end{document}